\documentclass[letterpaper, 10 pt, conference]{ieeeconf}  
\usepackage{graphics}
\usepackage{amsmath}
\usepackage{float}
\usepackage{graphicx}
\usepackage{tabularx}
\usepackage{comment}

\usepackage[font={small,it}]{caption}
\usepackage[font={small,it}]{subcaption}

\newcommand\Tstrut{\rule{0pt}{2.6ex}}         

\IEEEoverridecommandlockouts                              
\title{\LARGE \bf
Efficient Smoothing of Dilated Convolutions for Image Segmentation
}

\author{Thomas Ziegler, Manuel Fritsche, Lorenz Kuhn, Konstantin Donhauser\\ \{zieglert, manuelf, kuhnl, donhausk\}@ethz.ch
}

\begin{document}

\maketitle
\thispagestyle{empty}
\pagestyle{empty}

\begin{abstract}

 Dilated Convolutions have been shown to be highly useful for the task of image segmentation.
By introducing gaps into convolutional filters, they enable the use of larger receptive fields without increasing the original kernel size. Even though this allows for the inexpensive capturing of features at different scales, the structure of the dilated convolutional filter leads to a loss of information.

We hypothesise that inexpensive modifications to Dilated Convolutional Neural Networks, such as additional averaging layers, could overcome this limitation. In this project we test this hypothesis by evaluating the effect of these modifications for a state-of-the art image segmentation system and compare them to existing approaches with the same objective. 

Our experiments show that our proposed methods improve the performance of dilated convolutions for image segmentation. Crucially, our modifications achieve these results at a much lower computational cost than previous smoothing approaches.

\end{abstract}

\section{Introduction}

The goal in semantic image segmentation is to partition images and label each pixel in the resulting segments. Good segmentation algorithms are crucial for many real-world applications such as medical image processing \cite{Ronneberger2015UNet, litjens2017survey} or autonomous driving \cite{Ess2009SegmentationBasedUT, Geiger2012KITTI, cordts2016cityscapes, teichmann2018multinet}.

\begin{figure}
    \centering
    \includegraphics[width=1\linewidth]{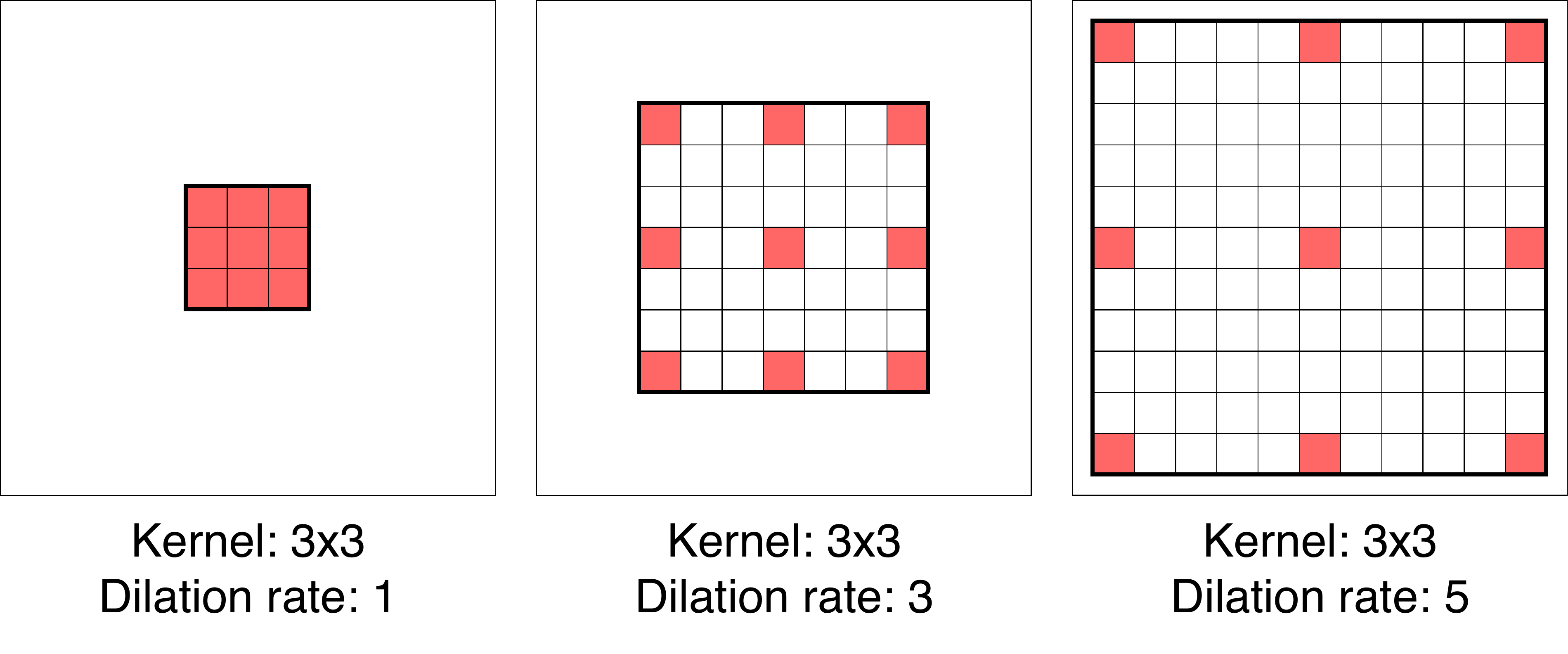}
    \caption{2D Dilated convolution with kernel size $3 \times 3$ and different dilation rates $r=1,3,5$. Dilated convolution enlarges the receptive field while keeping the number of parameters small.}
    \label{fig:dilated_kernel}
\end{figure}

One challenge in image segmentation is that objects may appear at different scales - both within the same image but also between images - which poses a problem for classical convolutional layers. The number of trainable parameters grows quadratically for increasing filter widths, making the use of larger convolutional filters prohibitively expensive. Learning features at different scales is thus difficult but essential to provide reliable image segmentation. DeepLab \cite{chen2018encoder} overcomes this issue by leveraging \textit{dilated convolutions}. These convolutions effectively introduce gaps into the filters, which increase the receptive field while maintaining the kernel size. In other words the filter size is increased but the number of weights stays the same. This is illustrated in figure \ref{fig:dilated_kernel}.

Dilated convolutions have proven to work well in practice and allow us to deal with object dependencies on different scales without reducing the image resolution. However, because of this sparse sampling only a few points are taken into account for potentially large parts of the image. If these points are noisy or simply bad representatives of their surroundings, the dilated convolution will yield bad results. Moreover, gridding artefacts occur, that is, adjacent pixels in the output are calculated from separate sets of pixels from the input \cite{wang2018understanding, wang2018smoothed}. This leads to a spatial information loss since neighbouring input pixels are usually related to each other. 

Recent methods \cite{wang2018understanding,wang2018smoothed} that were proposed to address these issues rely on introducing additional convolutional layers or stacking dilated convolution layers. While these methods are able to achieve small improvements in the segmentation quality, they effectively cancel out some of the benefits of using dilated convolutions. In particular, they overcome the information loss by adding a comparatively large number of trainable parameters to the dilated convolution, making the models more resource-intensive to train. 

To overcome these problems, we propose inexpensive modifications to dilated convolutions to make them more robust to local noise and encode more local spatial information. Rather than performing dilated convolutions directly on features, we first apply an additional interpolation filter on each input channel to capture more of the local information and then compute the dilated convolution on the filtered channels.

We use previously proposed methods to overcome the information loss \cite{wang2018smoothed} as baselines and show that that our modifications lead to networks which are significantly less expensive to train while achieving similar segmentation performance. 

In section \ref{sec:related_work}, we summarise the work on dilated convolutions for image segmentation as well as existing approaches for smoothing dilated convolutions. Section \ref{sec:methods} describes the problems of dilated convolution and our proposed modifications. Next, in section \ref{sec:results}  we introduce our experiments and explain the corresponding results. In section \ref{sec:discussion}, we discuss the benefit of our methods compare to the baselines. Lastly, in section \ref{sec:summary} we summarise our findings. Our implementation is publicly available on GitHub\footnote{https://github.com/ThomasZiegler/Efficient-Smoothing-of-Dilated-Convolutions}.

\section{Related Work}
\label{sec:related_work}
In this section, we outline previous work on semantic image segmentation and smoothing of dilated convolutions, which our contribution builds upon. 

DeepLab -- originally presented in \cite{chen2014semantic} and then improved upon in \cite{chen2018encoder, chen2018deeplab, chen2017rethinking} -- combines dilated convolutions with a number of other methods to achieve state-of-the-art performance on popular benchmark datasets \cite{cordts2016cityscapes, everingham2015pascal}. Multiple dilated convolution filters with different dilation rates are applied on the incoming filter map to cheaply and robustly segment objects at different scales.

Recent attempts to address the gridding artefacts \cite{wang2018understanding, hamaguchi2018effective}, visualised in figure \ref{fig:gridding}, have proposed strategies for selecting the dilation rate of consecutive dilated convolution layers which reduce the information loss.
In \cite{wang2018smoothed} additional convolutional layers are used to smooth the input of dilated convolution layers. The paper presents two methods, \textit{Smoothing by Group Interaction Layers (G Interact)} and \textit{Smoothing by Shared Separable Convolutions (SS Conv)}.
The central idea of \textit{G Interact} is to extend the network with additional layers after every dilated convolution layer. Each output of a the dilated convolution is then recomputed as linear combination of all outputs of the dilated convolutional layer. \textit{SS Conv}, on the other hand, adds an additional convolutional layer before every dilated convolution in the network. The same convolution is applied on each input channel separately. This can be seen as pre-filtering each channel with an fully learnable filter. These two methods achieve the state-of-the-art results in smoothing dilated convolutions for image segmentation. We thus use them as baselines for our own novel smoothing methods.
We note that \textit{G Interact} and \textit{SS Conv} add a non-trivial number of additional trainable parameters to each dilated convolution layer of a model which is already highly resource-intensive to train. 
Seeing this, we propose new smoothing methods which are not only conceptually simpler than the baseline methods, but are also less expensive to train and equally effective.

\section{Models and Methods}
\label{sec:methods}
\subsection{Dilated Convolutions}
Dilated convolutions have shown to be an effective way of capturing image features at different scales \cite{chen2018deeplab}. They introduce gaps into convolutional layers, which allows for an increased receptive field without introducing more weights. A filter with a dilation rate $r$ introduces $r\!-\!1$ zeros between the weights, which results in a sparse sampling of the input signal $x$. For a filter $w$ this can be written in 1D as
\begin{equation*}y[i]=\sum_{k=1}^K x[i-r\cdot k]w[k].\end{equation*}

Note that for a dilation rate $r\!=\!1$ this is equivalent to a standard convolution. An example for a 2D dilated convolution with kernel size $3\times3$ and different dilation rates is shown in figure \ref{fig:dilated_kernel}.

One issue with this sparse sampling is that only a few points are taken into account to calculate an output that is supposed to be dependent on a potentially large region of the image. These points can be bad representatives of their surroundings, if for example the image is noisy. 

\begin{figure}
    \centering
    \includegraphics[width=1\linewidth]{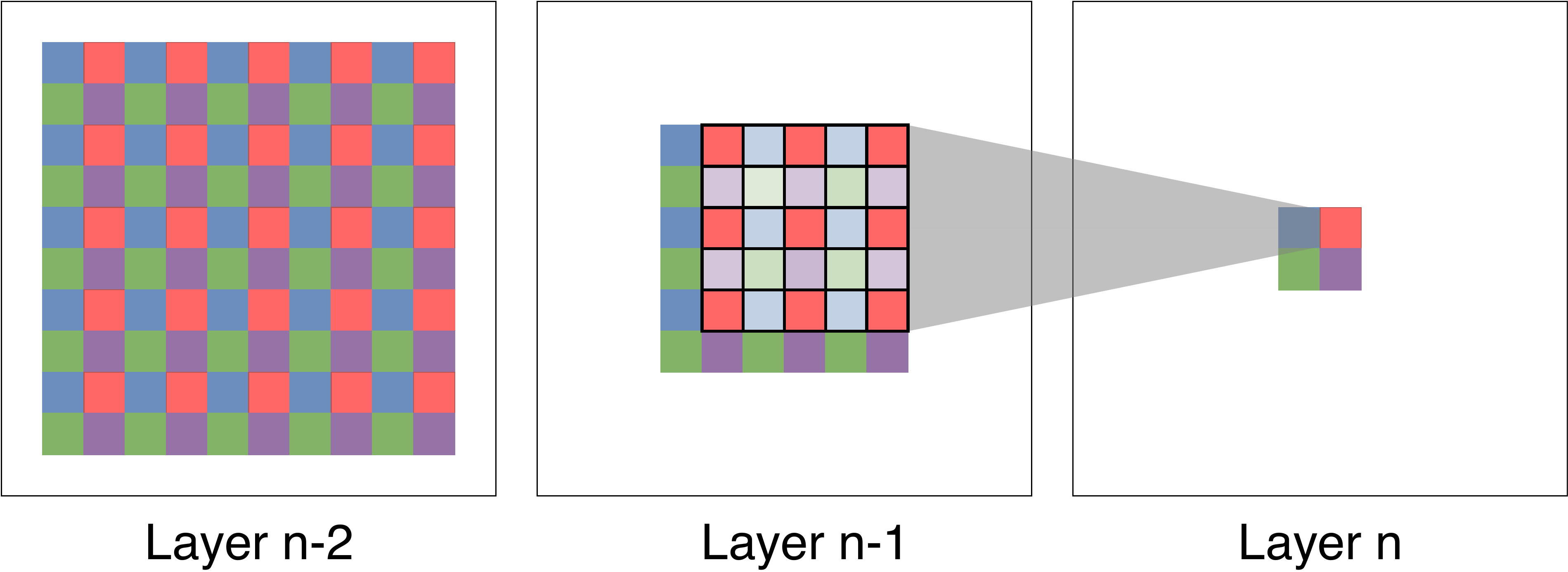}
    \caption{Visualisation of the gridding artefacts. Neighbouring pixels are not depending on the same input pixels if multiple dilated convolution layers are used. Here two layers of dilated convolutions with kernel size $3\times3$ and dilation rate $r=2$ are used. The coloured pixels in one layer $n$ are only dependant on the pixels of previous layers $(n\!-\!1$ and $n\!-\!2)$ with the same colour. The dilated convolution between layer $n$ and $n\!-\!1$ is shown explicit for the red pixel.}
    \label{fig:gridding}
\end{figure}

Another undesired result of dilated convolutions are so-called gridding artefacts \cite{wang2018understanding, wang2018smoothed}. Since the input is sampled sparsely with dilation rate $r$, for each output pixel the neighbouring $r$ pixels in every direction are computed without any shared input pixels. This results in a loss of spatial information, because these pixels can be completely independent of each other, even though they are spatially close together. This is visualised in figure~\ref{fig:gridding}. 

\subsection{Smoothed Dilated Convolution}
\begin{figure}
    \centering
    \includegraphics[width=0.49\linewidth]{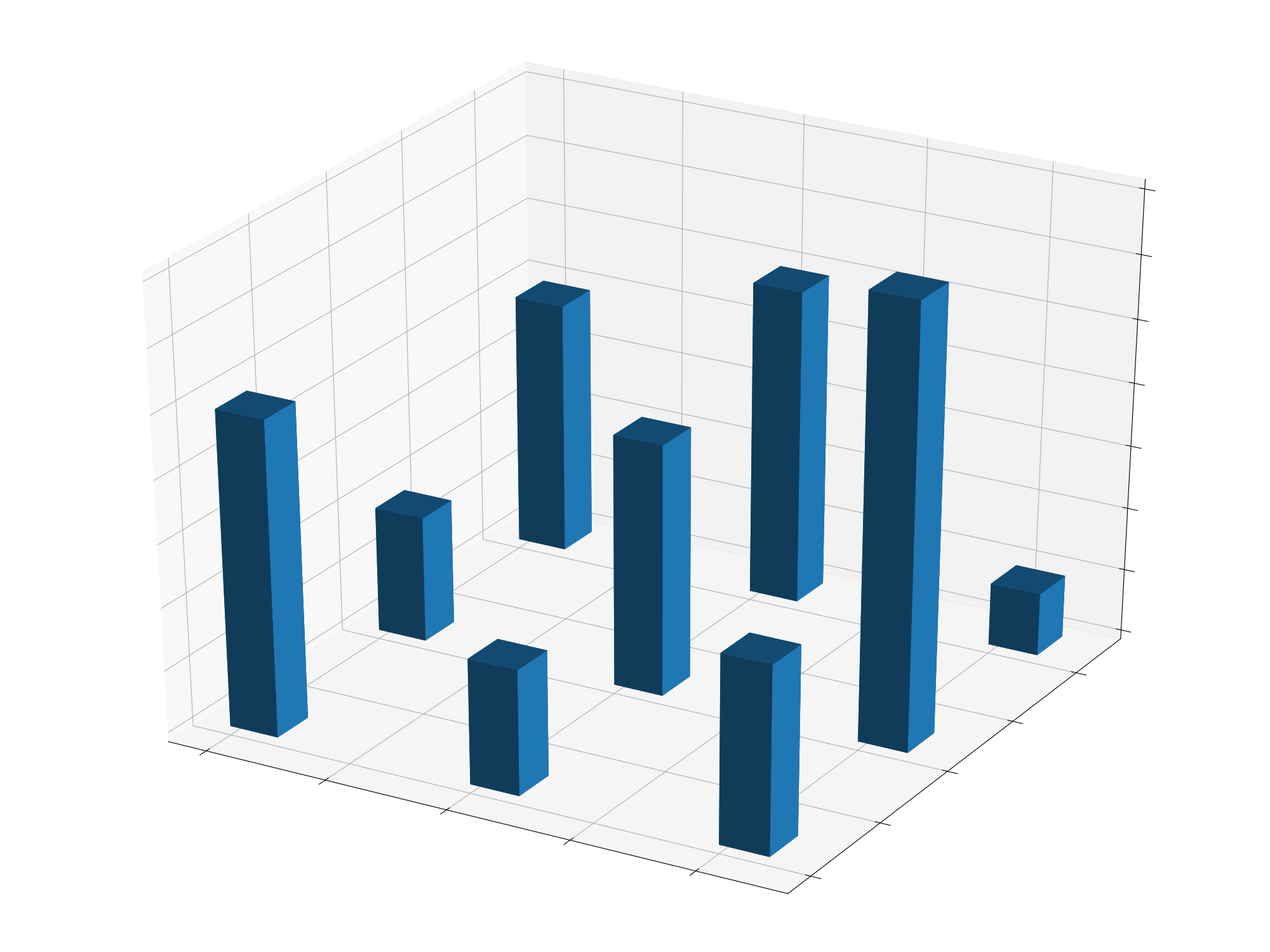}
    \includegraphics[width=0.49\linewidth]{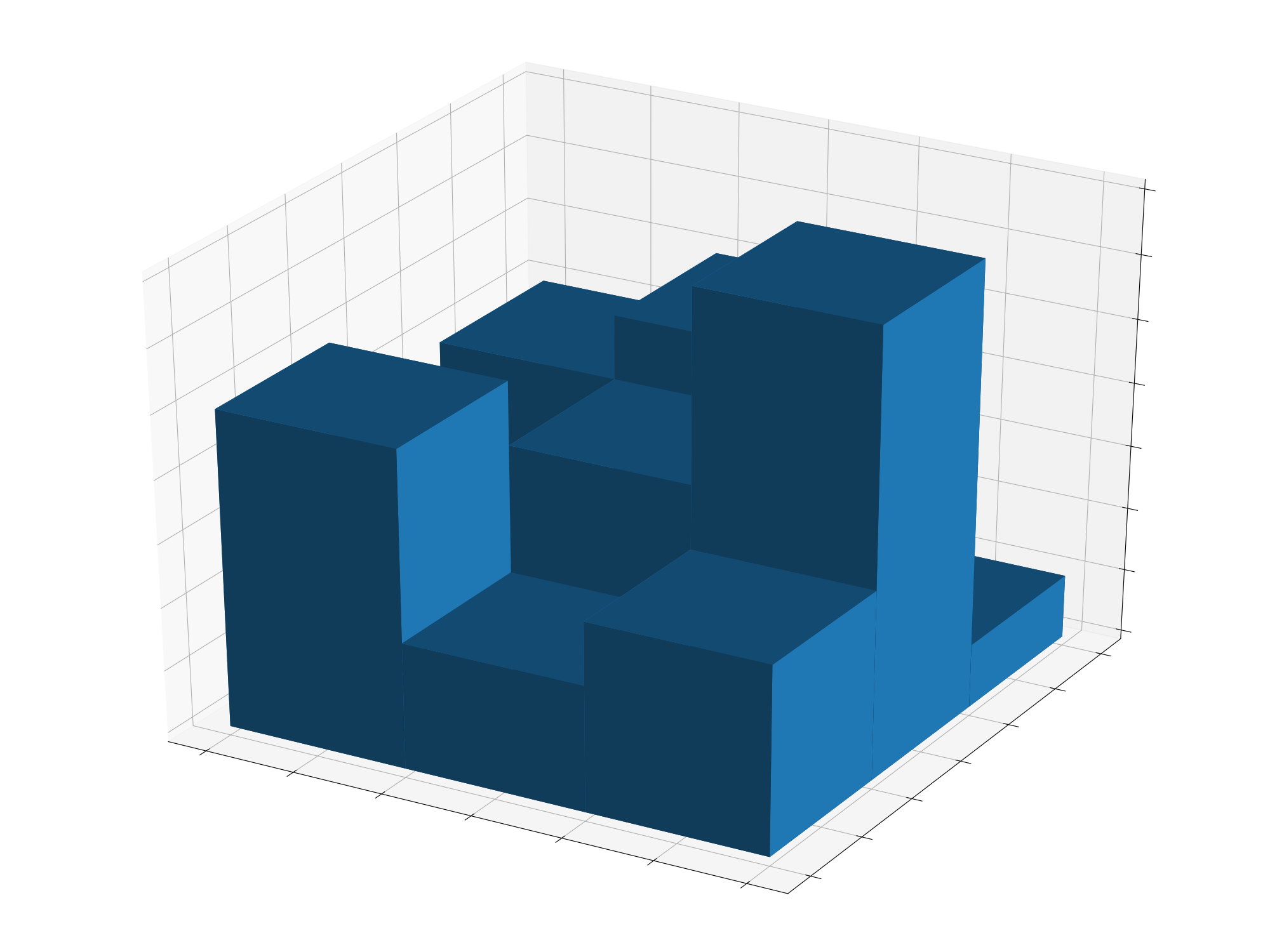}
    \caption{Left: An example of a dilated convolution with kernel size $3\times3$ and dilation rate $5$. Right: The smoothed version with a constant filter $v$ of size $r$. The added filter allows the dilated convolution to capture more local information at each of its weights.}
    \label{fig:conv_scheme}
\end{figure}

We propose the following method to overcome this problem: To encode more local information we combine neighbouring pixels in the input of the dilated convolutions. This can be achieved by using an additional interpolation filter $v$ with size $r$ before applying the dilated convolution. For each pixel this filter combines the information of surrounding pixels. Mathematically, this can be expressed in 1D as 
\begin{equation*}
y[i]=\sum_{k=1}^K\left(\sum_{n=-\lfloor r/2\rfloor}^{\lfloor r/2\rfloor} x[i-r\cdot k - n]v[n]\right)w[k].
\end{equation*}
Here it is assumed that $r$ is odd. In figure \ref{fig:conv_scheme} the two different convolution schemes are visualised for a constant filter $v$ of size $r$.

\subsection{Interpolation Methods}
\label{sec:Interpolation_Methods}
\subsubsection{Average}
A common down-sampling approach is to calculate the bilinear interpolation of pixels. Inspired by this, our first choice for the input filter is to calculate the average of the surrounding points. The resulting combination of input filter and dilated convolution is visualised in figure \ref{fig:conv_scheme} (right) and can be written as
\begin{equation*}
    v[x, y] = \begin{cases} \frac{1}{r^2} & \text{, if }\lvert x \rvert, \lvert y \rvert  < \lfloor r/2 \rfloor\\
    0 & \text{, otherwise.}
    \end{cases}
\end{equation*}

\subsubsection{Gaussian}
\label{subsec:Gaussian}
Applying the averaging filter to an input assigns the same weight to a given pixel's neighbours and zero weight to pixels further away. While this works, it does not adequately capture the intuition that a pixel's immediate neighbours should influence the dilated convolution's output more strongly than more remote pixels. We thus introduce a second filter which puts more weight on pixels close to the center of the filter and less weight on ones that are close to the edge of the filter. A common function that satisfies these requirements is the Gaussian with variance $\sigma^2$:
\begin{equation*}
    v[x, y] = \begin{cases} \frac{1}{2\pi \sigma^2}\exp{\left(-\frac{1}{2\sigma^2}(x^2+y^2)\right)} & \text{, if }\lvert x \rvert, \lvert y \rvert  < \lfloor r/2 \rfloor\\
    0 & \text{, otherwise.}
    \end{cases}
\end{equation*}
Here, $\sigma$ is a fixed parameter that can be selected empirically.

\subsubsection{Trainable}
Finally, the parameters for the input filter $v$ can also be learned, which corresponds to baseline \textit{SS Conv} method \cite{wang2018smoothed} described above. While this allows the network to optimise its input filter, it also introduces a large number of trainable weights.

Our proposed interpolation filters can be implemented as a depthwise separable convolution \cite{Chollet2016SeparableConv}. 
    
\subsection{Aggregated Filters}
\label{subsec:aggregation}
We investigate the performances of the methods described in section \ref{sec:Interpolation_Methods}. The most obvious way to do this is to run each of them and then compare the results. However, one can also include a convex combination of the different filters in the network, and then let the network optimise on what filters it uses. This also gives some insights on which method works best. To do that we use all the filters described in section \ref{sec:Interpolation_Methods} in parallel, which we denote as $v_{\text{Ave}}$, $v_{\text{Gauss}}$ and $v_{\text{Learned} }$ in the following. Additionally, we also add a filter with the same input and output, meaning no filtering is performed, which we denote as $\delta$. We combine them with a simple convex combination: 
\begin{equation*}
v = \alpha_{\text{Ave}}v_{\text{Ave}} + \alpha_{\text{Gauss}}v_{\text{Gauss}} + \alpha_{\text{Learned}}v_{\text{Learned}} + \alpha_{\text{None}}\delta,
\end{equation*}
\begin{equation*}
\text{where} \quad \alpha_{\text{Ave}} + \alpha_{\text{Gauss}} + \alpha_{\text{Learned}} + \alpha_{\text{None}} = 1, \quad \alpha_i\geq 0.
\end{equation*}
The $\alpha_i$ coefficients are learned during training. This combined filter is then used as the input to the dilated convolution. The idea is that -- after training the whole network -- we get some additional insights on which filter was ``choosen" by the optimiser, by comparing the different coefficients $\alpha_i$.

\section{Results}
\label{sec:results}

\subsection{Datasets}
We evaluate our methods on the PASCAL VOC 2012 \cite{everingham2015pascal} and Cityscapes \cite{cordts2016cityscapes} datasets. Both of them contain different object classes and provide images with pixel-wise annotations as labels. The performance on these datasets is measured in terms of pixel Intersetion-over-Union (IoU) on the different classes and the mean IoU (mIoU) over all classes.

\subsubsection*{PASCAL VOC 2012}
This dataset contains 20 object classes and one background class. The original dataset is divided into \textit{train}, \textit{val} and \textit{test} sets with $1\,464$, $1\,449$, and $1\,456$ images, respectively. We use the augmented version \cite{Hariharan2011Semantic} which provides extra annotations, increasing the size of the \textit{train} set to a total of $10\,582$ images. The models are trained on randomly cropped patches of size $321 \! \times  \!321$ from the augmented \textit{train} set. The validation is performed using the \textit{val} set.

\subsubsection*{Cityscapes}
This dataset shows street scenes from various cities in Germany and Switzerland with annotations from 30 different object classes. As is typically done, we ignore void categories and rare cases and only use 19 classes ignoring the rest of them. The dataset is divided into \textit{train}, \textit{val} and \textit{test}, with $2\,975, 500$ and $1\,525$ images, respectively. The models are trained on randomly cropped patches of size $721\!\times \!721$ from the \textit{train} set and evaluated on the \textit{val} set.

\begin{table*}[!t]
	\centering
	\begin{tabular*}{\linewidth}{l|l|c*{21}{p{0.015\hsize}}}
		\textbf{System} \Tstrut & \textbf{mIoU} & \textbf{1} & \textbf{2} & \textbf{3} & \textbf{4} & \textbf{5} & \textbf{6} & \textbf{7}  & \textbf{8} & \textbf{9} & \textbf{10} & \textbf{11} & \textbf{12} & \textbf{13} & \textbf{14} & \textbf{15} & \textbf{16} & \textbf{17} & \textbf{18} & \textbf{19} & \textbf{20} & \textbf{21} \\ 
		\hline
		DeepLabv2 \Tstrut & 72.0 & 93.0 & 83.3 & 39.0 & 82.7 & 63.7 & 72.1 & 91.2 & 81.9 & 86.5 & 35.6 & \textbf{76.5} & 54.8 & 80.7 & 77.9 & 78.0 & 83.4 & 56.3 & 81.0 & 40.2 & 81.5 & 73.1  \\	
		SS Conv & 72.5 & \textbf{93.2} & 84.5 & 38.8 & 83.6 & \textbf{67.2} & 70.8 & \textbf{92.5} & \textbf{84.9} & 88.5 & 34.0 & 74.7 & 52.4 & 80.6 & 81.5 & 80.3 & \textbf{84.1} & 54.9 & 79.8 & 43.3 & \textbf{81.8} & 72.4 \\
		G Interact & \textbf{73.0} & \textbf{93.2} & \textbf{84.9} & 38.6 & \textbf{84.8} & 65.7 & 74.2 & \textbf{92.5} & 84.5 & \textbf{88.7} & \textbf{37.1} & 75.3 & 57.4 & \textbf{82.7} & 78.6 & 80.1 & 83.6 & \textbf{56.6} & 79.4 & 41.8 & 80.1 & \textbf{73.7} \\  
		\hline
		Average \Tstrut & 72.7 & 93.1 & 84.6 & \textbf{39.4} & 84.5 & 64.8 & 74.3 & 91.8 & 84.1 & 88.3 & 35.2 & 75.3 & \textbf{55.1} & 81.6 & 80.6 & 79.5 & 83.2 & 55.8 & \textbf{81.9} & 41.5 & 80.5 & 72.5 \\  
		Gaussian & 72.9 & 93.1 & 84.4 & 39.2 & 82.3 & 65.7 & \textbf{76.0} & 92.1 & 84.6 & 88.3 & 35.7 & 75.8 & 55.0 & 81.1 & \textbf{81.9} & \textbf{80.8} & 83.2 & 54.9 & 79.7 & \textbf{43.9} & 81.6 & 72.4 \\  
	\end{tabular*}
	\caption{IoU scores of the different classes and the mean IoU (mIoU) over all classes in \% of our novel methods as well as the baseline methods on the PASCAL VOC 2012 dataset. Best score per class is marked bold. Class 1 is background and class 2-19 represent: aeroplane, bicycle, bird, boat, bottle, bus, car, cat, chair, cow, diningtable, got, horse, motorbike, person, potteplant, sheep, sofa, train tvmonitor".}
	\label{table:voc-results}
\end{table*}

\begin{figure*}[!t]
	\centering
	\begin{subfigure}[t]{0.135\linewidth}  
		\caption{Image}
		\includegraphics[width=1\linewidth]{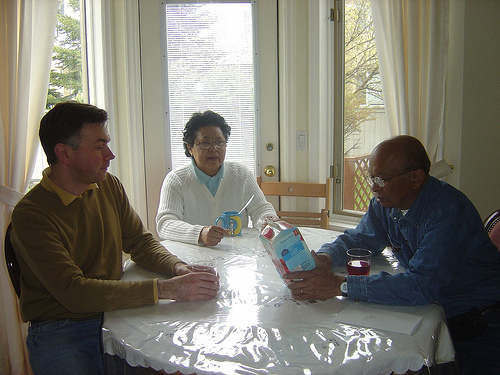}
	\end{subfigure}
	\begin{subfigure}[t]{0.135\linewidth}  
		\caption{Ground Truth}
		\includegraphics[width=1\linewidth]{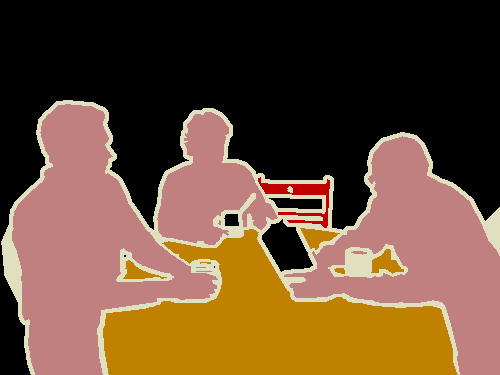}
	\end{subfigure}
	\begin{subfigure}[t]{0.135\linewidth}  
		\caption{DeepLabv2}
		\includegraphics[width=1\linewidth]{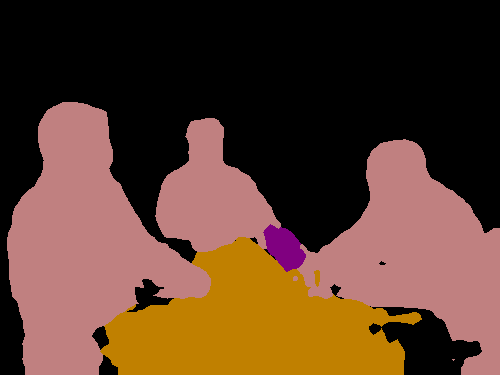}
	\end{subfigure}
	\begin{subfigure}[t]{0.135\linewidth}  
		\caption{SS Conv}
		\includegraphics[width=1\linewidth]{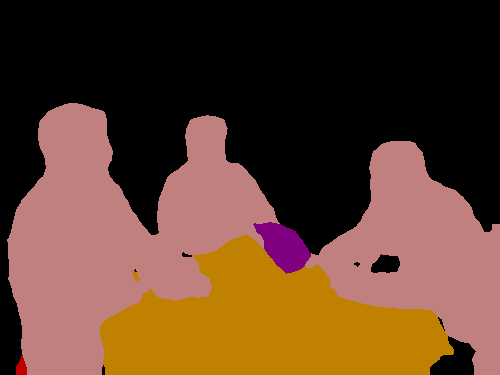}
	\end{subfigure}
	\begin{subfigure}[t]{0.135\linewidth}  
		\caption{G Interact}
		\includegraphics[width=1\linewidth]{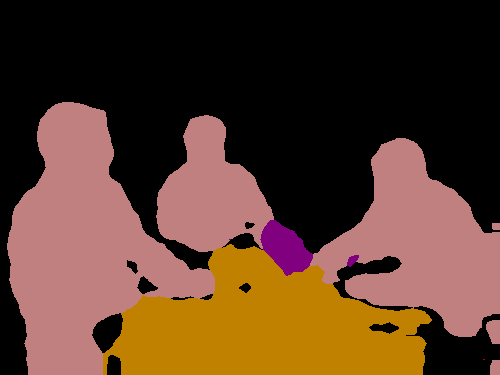}
	\end{subfigure}
	\begin{subfigure}[t]{0.135\linewidth}  
		\caption{Average}
		\includegraphics[width=1\linewidth]{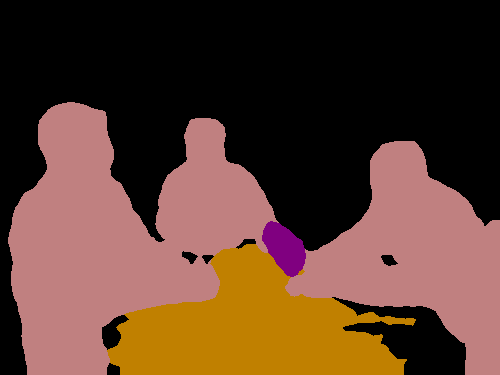}
	\end{subfigure}
	\begin{subfigure}[t]{0.135\linewidth}  
		\caption{Gaussian}
		\includegraphics[width=1\linewidth]{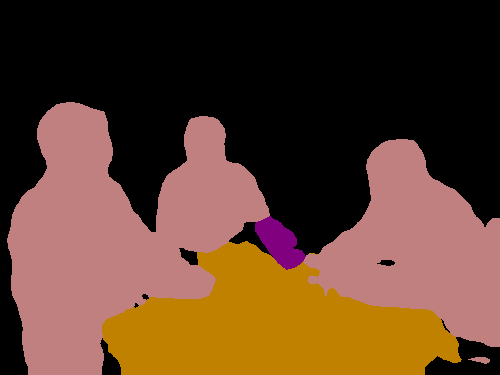}
	\end{subfigure}
	
	\vspace{0.1cm}
	
	\includegraphics[width=0.135\linewidth]{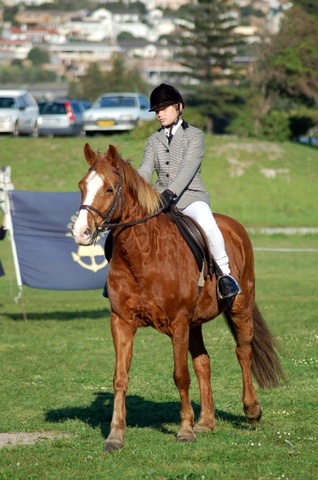}
	\includegraphics[width=0.135\linewidth]{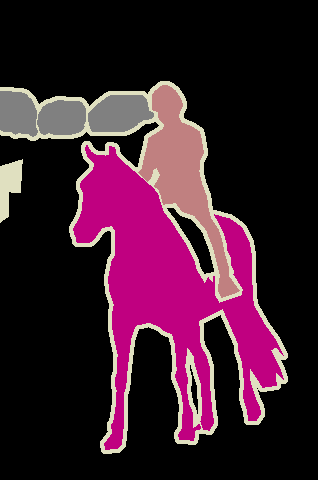}
	\includegraphics[width=0.135\linewidth]{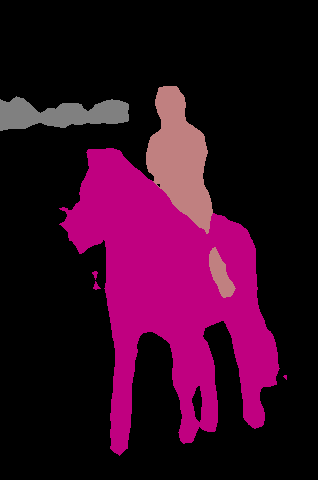}  
	\includegraphics[width=0.135\linewidth]{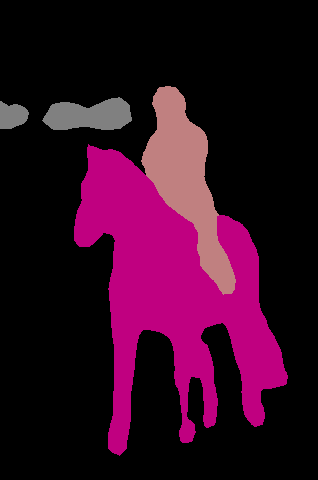}
	\includegraphics[width=0.135\linewidth]{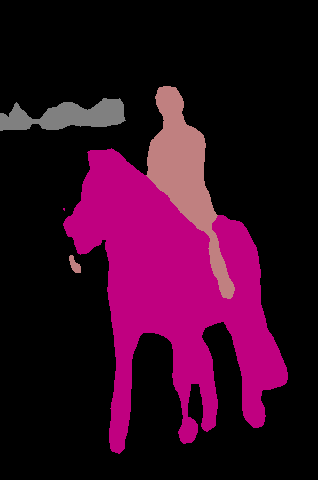}
	\includegraphics[width=0.135\linewidth]{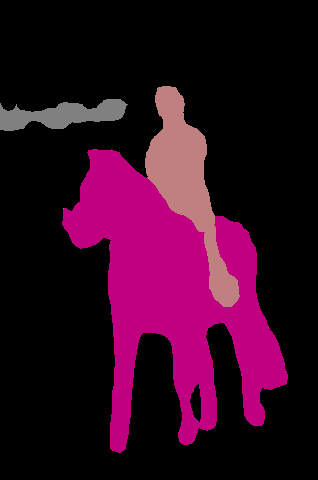}
	\includegraphics[width=0.135\linewidth]{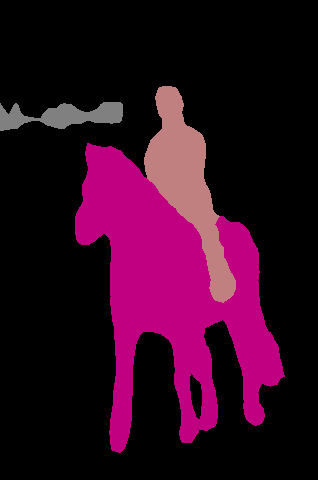}
	\caption{Segmentation examples of the different methods on the PASCAL VOC 2012 dataset.}  
	\label{fig:voc_images}
\end{figure*}

\subsection{Individual Filters on PASCAL VOC 2012}
\label{experiment:VOC}
As baselines we compare our methods to DeepLabv2 \cite{chen2018deeplab} and the two smoothing approaches presented in \cite{wang2018smoothed} that were built on DeepLabv2. In all cases we do not use any post-processing steps such as conditional-random-fields \cite{chen2018deeplab}. Another method for improving dilated convolutions is presented in \cite{wang2018understanding}. However, they use a different version of DeepLab, which is why the results are not comparable and we do not include them here. Our implementation builds upon the Tensorflow reimplementation of DeepLabv2 provided by \cite{wang2018smoothed}, which also includes their two proposed smoothing approaches. For all runs we use the hyper-parameters from \cite{wang2018smoothed}, except for the initial learning rate. A detailed description of the model parameters can be found in their work. The initial learning rate is chosen as $0.002$. The batch size is set to $10$ during training. All methods are trained for $20\,000$ steps. The $\sigma$ parameter in the \textit{Gaussian} filter is set to $1.0$, which we found by line search. All runs were performed on a single GTX 1080Ti GPU. We restricted our experiments to the version of DeepLabv2 which is pre-trained on the MS-COCO dataset \cite{Lin2014MSCOCO}.

As can be seen in table \ref{table:voc-results}, our methods \textit{Average} and \textit{Gaussian} improve the IoU for most classes as well as the mean IoU (mIoU) compared to baseline DeepLabv2. They achieve comparable segmentation results as the baseline methods \textit{SS Conv} and \textit{G Interact} for the different classes and the mean over all classes. Crucially, our methods are able to achieve this performance while being significantly more efficient to train than the baseline smoothing methods. In table \ref{table:voc-runtimes} we list the time it takes to perform $20\,000$ update steps for all evaluated models. While \textit{G Interact} and \textit{SS Conv} increase the training time of the system by $13.8\%$ and $24.9\%$, respectively, our methods only increase the time by $4.3\%$ and $5.6\%$ respectively. This means that our proposed methods achieve a similar gain in segmentation quality as previous smoothing method while being roughly three times more time-efficient on this dataset. 
In figure \ref{fig:voc_images} some example segmentation results are shown. In these examples \textit{SS Conv} and \textit{Gaussian} yield similar segmentations, which are closest to the ground truth compared to the other methods. As expected, the pure DeepLabv2 architecture produces poorer results than the other methods.

\begin{table}[]
	\resizebox{\linewidth}{!}{%
		\begin{tabular}{l|lllll}
			
			\textbf{} & \textbf{DeepLabv2} & \textbf{SS Conv} & \textbf{G Interact} & \textbf{Average} & \textbf{Gaussian} \\ \hline
			\Tstrut \textbf{20k Updates} \Tstrut & 9h 10min & 11h 27min & 10h 26min &9h 34 min & 9h 41min \\ 
			\textbf{Add. runtime} & 0\% & 24.9\% & 13.8\% & 4.3\% & 5.6\% \\ 
		\end{tabular}%
	}
	\caption{Training time for $20\,000$ update steps on the PASCAL VOC 2012 dataset for our novel methods as well as the baseline methods. The additional time compared to the pure DeepLabv2 system is also provided.}
	\label{table:voc-runtimes}
\end{table}

\begin{figure}
	\centering
	\includegraphics[width=1\linewidth]{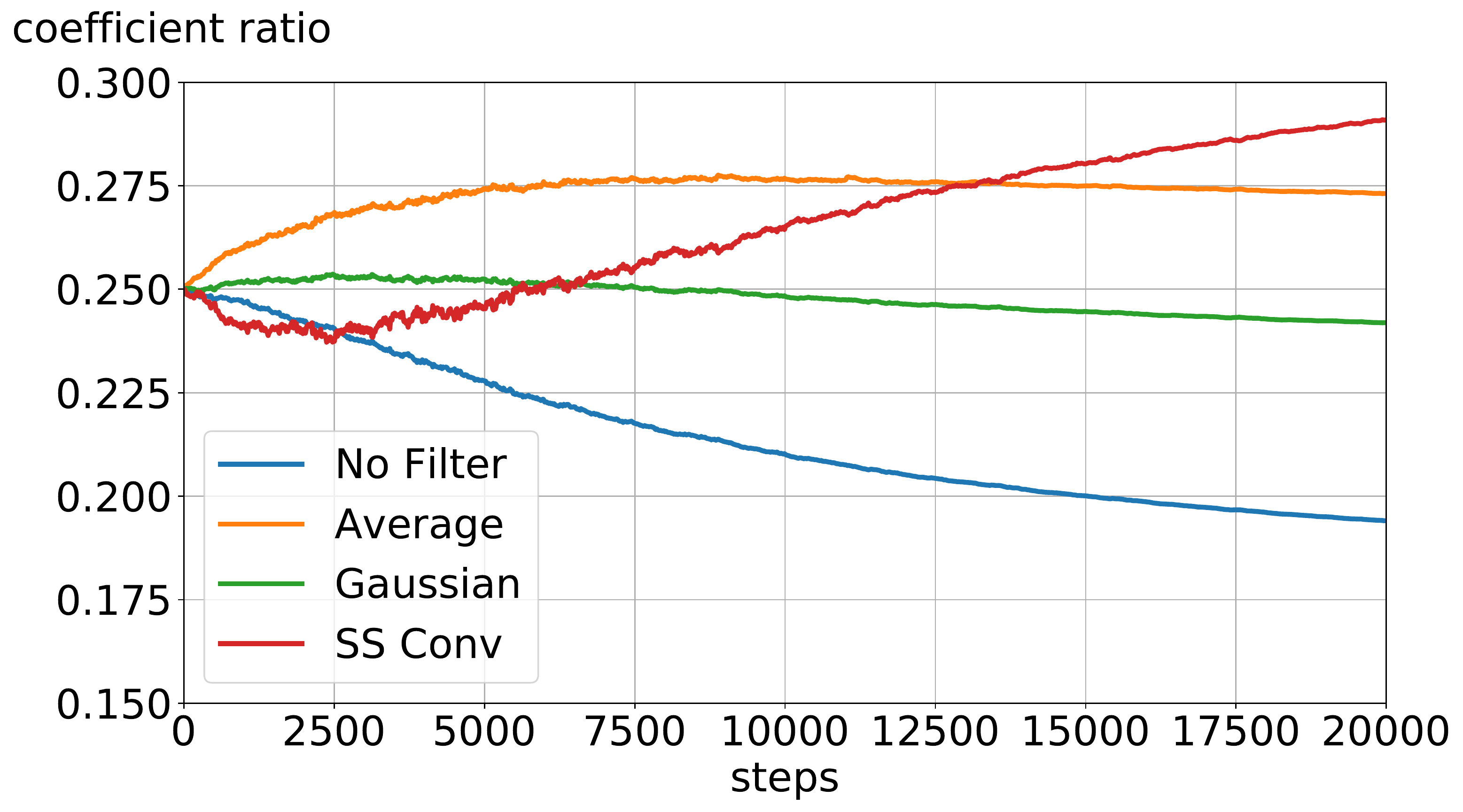}
	\caption{Coefficient ratio of the convex combination of the proposed interpolation filters over $20\,000$ training steps on the PASCAL VOC 2012 dataset.}
	\label{fig:aggregation}
\end{figure}

\begin{table*}[htb]
	\centering
	\begin{tabular*}{\linewidth}{l|l|c*{19}{p{0.019\hsize}}}
		\textbf{System} \Tstrut & \textbf{mIoU} & \textbf{1} & \textbf{2} & \textbf{3} & \textbf{4} & \textbf{5} & \textbf{6} & \textbf{7}  & \textbf{8} & \textbf{9} & \textbf{10} & \textbf{11} & \textbf{12} & \textbf{13} & \textbf{14} & \textbf{15} & \textbf{16} & \textbf{17} & \textbf{18} & \textbf{19} \\ 
		\hline
		DeepLabv2 \Tstrut & 68.07 & \textbf{96.6} & \textbf{74.5} & \textbf{89.4} & 37.1 & 45.6 & 49.9 & 53.6 & 65.5 & \textbf{89.8} & \textbf{53.7} & 92.6 & 74.9 & \textbf{51.8} & \textbf{92.3} & 63.9 & \textbf{75.8} & 60.5 & 55.8 & \textbf{70.6}   \\	
		SS Conv & 68.19 & \textbf{96.6} & \textbf{74.5} & \textbf{89.4} & \textbf{38.6} & \textbf{45.8} & 49.5 & 54.1 & 65.7 & \textbf{89.8} & 52.5 & 92.6 & \textbf{75.0} & 50.2 & 92.2 & 64.2 & 75.5 & 63.7 & 55.4 & 70.2  \\
  
		\hline
		Average \Tstrut & \textbf{68.36} & \textbf{96.6} & \textbf{74.5} & 89.3 & 36.2 & 44.3 & \textbf{50.1} & \textbf{54.5} & \textbf{66.1} & 89.7 & 52.9 & \textbf{92.8} & 74.9 & 50.9 & 92.2 & \textbf{64.4} & \textbf{75.8} & \textbf{64.0} & 56.8 & \textbf{70.6} \\  
		Gaussian & 68.12 & 96.5 & 74.0 & 89.3 & 38.0 & 45.3 & 50.0 & 53.5 & 65.3 & \textbf{89.8} & 53.6 & 92.6 & \textbf{75.0} & 51.2 & \textbf{92.3} & \textbf{64.4} & 74.3 & 61.0 & \textbf{57.7} & 70.4 \\  
	\end{tabular*}
\caption{IoU scores of the different classes and the mean IoU (mIoU) over all classes in \% of our novel methods as well as the baseline methods on the Cityscapes dataset. Best score per class is marked bold. The class 1-19 represent: road, sidewalk, building, wall, fence, pole, traffic light, traffic sign, vegetation, terrain, sky, person, rider, car, truck, bus, train, motorcycle, bicycle.}
	\label{table:cityscapes-results}
\end{table*}

\subsection{Filter Aggregation on PASCAL VOC 2012}
\label{experiment:Aggregation}
As described in section \ref{subsec:aggregation}, another method to evaluate the performance of the different smoothing filters is to use a convex combination of the filters and let the system learn the coefficients. We analyse this on the PASCAL VOC 2012 dataset. The used smoothing filters are \textit{Average}, \textit{Gaussian}, \textit{SS Conv} and \textit{no Filter}. The first two correspond to our proposed methods. The third is one of the proposed methods from \cite{wang2018smoothed} and the last one can be seen base version of Deeplabv2.

Combining all filters requires noticeably more GPU memory than a single filter. Hence, we run this experiment with a reduced batch size of $9$, all other hyperparameter values are the same as described in section \ref{experiment:VOC}.

Figure \ref{fig:aggregation} shows the coefficients of the convex combination over $20\,000$ learning steps. The coefficient for non filtering decreases from the beginning, showing that any smoothing method brings advantage. Furthermore one can see that \textit{SS Conv}'s weight decreases at the beginning since the filter is randomly initialised. As the learning progresses, however, its weight increases steadily. This might indicate that allowing individually learnable filter weights may produce the best segmentation results given enough resources. On the other hand, this experiment does not exactly correspond to the individual performance of the filters since it is per-definition a combination of them and none of the coefficients goes to 0. The \textit{Average} filter seems to outperform the \textit{Gaussian} filter according to their weights in the convex combination which is not the result we observe when comparing the filters individually as in section \ref{experiment:VOC}.


\subsection{Individual Filters on Cityscapes}
\label{experiment:Cityscapes}
On the Cityscapes dataset we used the same baselines as in section \ref{experiment:VOC} with some changes in the parameters due to the increased image sizes. We change the batch size to $3$ and the initial learning rate to $0.0001$. The maximum number of training steps is also increased to $80\,000$ to ensure convergence. These hyperparameters are used for all methods in this experiment. 

The results on \textit{G Interact} are omitted due to GPU memory limitations. To successfully run \textit{G Interact} the batch size would have to be reduced to $2$, yielding results not comparable to the other runs. 

As shown in table \ref{table:cityscapes-results}, our method \textit{Gaussian} achieves a marginal improvement in the overall segmentation quality (mIoU) compared to the baseline DeepLabv2, whereas \textit{Average} achieves a larger improvement and also outperforms the baseline method \textit{SS Conv}.

\begin{table}[]
\resizebox{\linewidth}{!}{%
\begin{tabular}{l|llll}

\textbf{} & \textbf{DeepLabv2} & \textbf{SS Conv} &  \textbf{Average} & \textbf{Gaussian} \\ \hline
\textbf{Training Steps}\Tstrut & $68\,000$ & $64\,000$ & $70\,000$ & $68\,000$ \\
\textbf{Training Time} & 59h 53min & 73h 05min & 62h 29min & 61h 37min  \\ 
\textbf{Add. runtime} & 0\% & 22.04 \% & 4.34\% & 2.89\% \\ 
\end{tabular}%
}
\caption{The number of steps and corresponding training time until maximum evaluation result is achieved on the Cityscapes dataset. The additional time compared to the naive DeepLabv2 system is also provided.}
	\label{table:cityscapes-runtimes}
\end{table}

In table \ref{table:cityscapes-runtimes}, we compare the number of training steps and the time it takes to achieve the best evaluation result for the different methods. Our \textit{Gaussian} filter needs the same number of training steps as the pure Deeplabv2 ($68\,000$) and our \textit{Average} an additional $2\,000$ steps. They have a slight increase in training time of  $+2.89\%$ and $+4.34\%$, respectively. \textit{SS Conv} achieves its results in less training steps ($64\,000$). However, its training time is still significantly longer than all other methods ($+\!22.04\%$) due to the additional trainable weights.

\section{Discussion}
\label{sec:discussion}
This work started off from the hypothesis that the issues linked to dilated convolutions for image segmentation could be addressed using conceptually simpler methods than those proposed in the existing literature. This project makes two important contributions in this regard. Firstly, our results suggest that relatively simple smoothing methods can indeed achieve similar improvements of segmentation quality as the complex approaches discussed in section \ref{sec:related_work}. We achieved a performance improvement on both the PASCAL VOC 2012 and Cityscapes datasets, indicating that our filters are not dataset specific. Secondly, our work shifts the focus from considering only segmentation scores to including the cost at which these improvements are achieved. We do this by making the training time of the evaluated systems a criterion of their quality and point out the potentially large computational penalty for relatively small improvements of segmentation performance. 

At the same time, there are a number of limitations in our work. Having limited computational resources, we were not able to repeatedly run the experiments using different random seeds and perform an extensive parameter optimisation as would be best practice. 
For the same reasons, we were not able to do a parameter optimisation for each individual model. Nonetheless, we show that even using non-optimised hyperparameters, our models perform comparably well. We also were not able to compare our results for the baseline methods to published results since we had to use slightly modified configurations parameters to avoid numerical instabilities with our novel filters. However, we don't expect this to affect the validity of our results.

\section{Summary}
\label{sec:summary}
In this paper we introduced an extension to dilated convolutions which allows them to encode more spatial information. To achieve that we apply simple additional filters to the input signal of these convolutions. These filters are easy to implement and introduce no additional parameters to train. Despite their simplicity, these modifications achieve similar performance gains as the current state-of-the-art smoothing methods, while requiring significantly less time to train. Thus, our results suggest that our methods are a good alternative to existing methods to overcome the issues of dilated convolutions.

\addtolength{\textheight}{-12cm}   



\bibliographystyle{ieeetr}
\bibliography{refs.bib}

\end{document}